\documentclass[letterpaper, 10 pt, conference]{ieeeconf}  

\IEEEoverridecommandlockouts                              

\usepackage{cite}
\usepackage{graphicx}
\usepackage{amsmath,amssymb}
\usepackage{algorithm}
\usepackage{algpseudocode}
\usepackage{booktabs}
\usepackage{xcolor}
\usepackage[hidelinks]{hyperref}
\usepackage{siunitx}
\usepackage{cleveref}

\newcommand{\f}{f}

\newcommand{\tabnote}[2][\columnwidth]{%
  \par\vspace{2pt}\parbox{#1}{\scriptsize #2}}
\newcommand{\cmt}[1]{\Statex\textcolor[gray]{0.45}{\texttt{\footnotesize //~#1}}}
\algrenewcommand\algorithmiccomment[1]{\hfill\textcolor[gray]{0.45}{\texttt{\footnotesize //~#1}}}
\newcommand{\InlineComment}[1]{%
  \unskip\nobreak\hfill
  \makebox[0pt][r]{%
    \textcolor[gray]{0.45}{\texttt{\tiny //~#1}}%
  }%
}

\title{\LARGE \bf
Steering Generative Robot Policies with Lexicographic Preferences
}

\author{Yixuan Jia and Jonathan P.\ How%
\thanks{Y.\ Jia and J.\ How are with Department of Aeronautics and Astronautics, Massachusetts Institute of Technology, Cambridge, MA 02319 USA (e-mail: yixuany@mit.edu, jhow@mit.edu).}%
}

\begin{document}

\maketitle
\thispagestyle{empty}
\pagestyle{empty}

\begin{abstract}



Pretrained generative robot policies can produce effective behaviors across diverse
environments, but deployment can lead to requirements and preferences that may not have been
represented during training. Furthermore, at deployment, an operator, user, or application may assign
these requirements and preferences a priority order that can vary across deployments.
For example, embodiment-specific feasibility constraints may need to be
satisfied first, while user-specific preferences guide behavior among the feasible options.
We show that a frozen generative robot policy---based on either diffusion or flow
matching---can be steered at \emph{inference time} to respect such lexicographically ordered
deployment objectives.

To achieve this, we introduce two modifications to the sampler. First, we apply
dynamic-barrier guidance to sampled trajectories, constraining lower-priority updates so
that higher-priority costs do not increase (up to first order). Second, we select the
executed sample using a cascade that successively filters candidate samples according to
each priority level. The policy weights remain unchanged.

On a navigation benchmark, we demonstrate that our method improves success, traversability, and preference compliance over the frozen policy, and achieves substantially better compliance than tuned weighted-sum baselines. The same method transfers to a flow-matching manipulation policy on LIBERO, where it improves compliance without reducing task success. A controlled manipulation study further shows that, in settings where a fixed weight can match the desired ordering, the dynamic barrier reaches comparable best performance over a substantially wider range of parameter settings.
\end{abstract}

\section{INTRODUCTION}
Large pretrained generative robot policies can map raw sensory input directly to motion without a map, a planner, or task-specific engineering, and their capabilities continue to improve~\cite{chi2025diffusion,shah2023vint,sridhar2024nomad,cai2025navdp}. However, broad generalization does not eliminate the need to adapt behavior to the requirements of a particular deployment. These requirements may include embodiment-specific feasibility constraints as well as user- or application-specific preferences that were not represented during training. A navigation policy, for example, may propose a route that is not traversable by the particular robot on which it is deployed, or a delivery robot’s operator may prefer that it not cut across residential lawns. Similarly, a household may prefer that a manipulator not carry a dripping bowl over the stove. Such behaviors need not reflect a failure of the pretrained policy itself; rather, they expose a gap between the broad behaviors learned during pretraining and the specific requirements imposed at deployment.

\begin{figure}[t]
\centering
\includegraphics[width=0.49\textwidth]{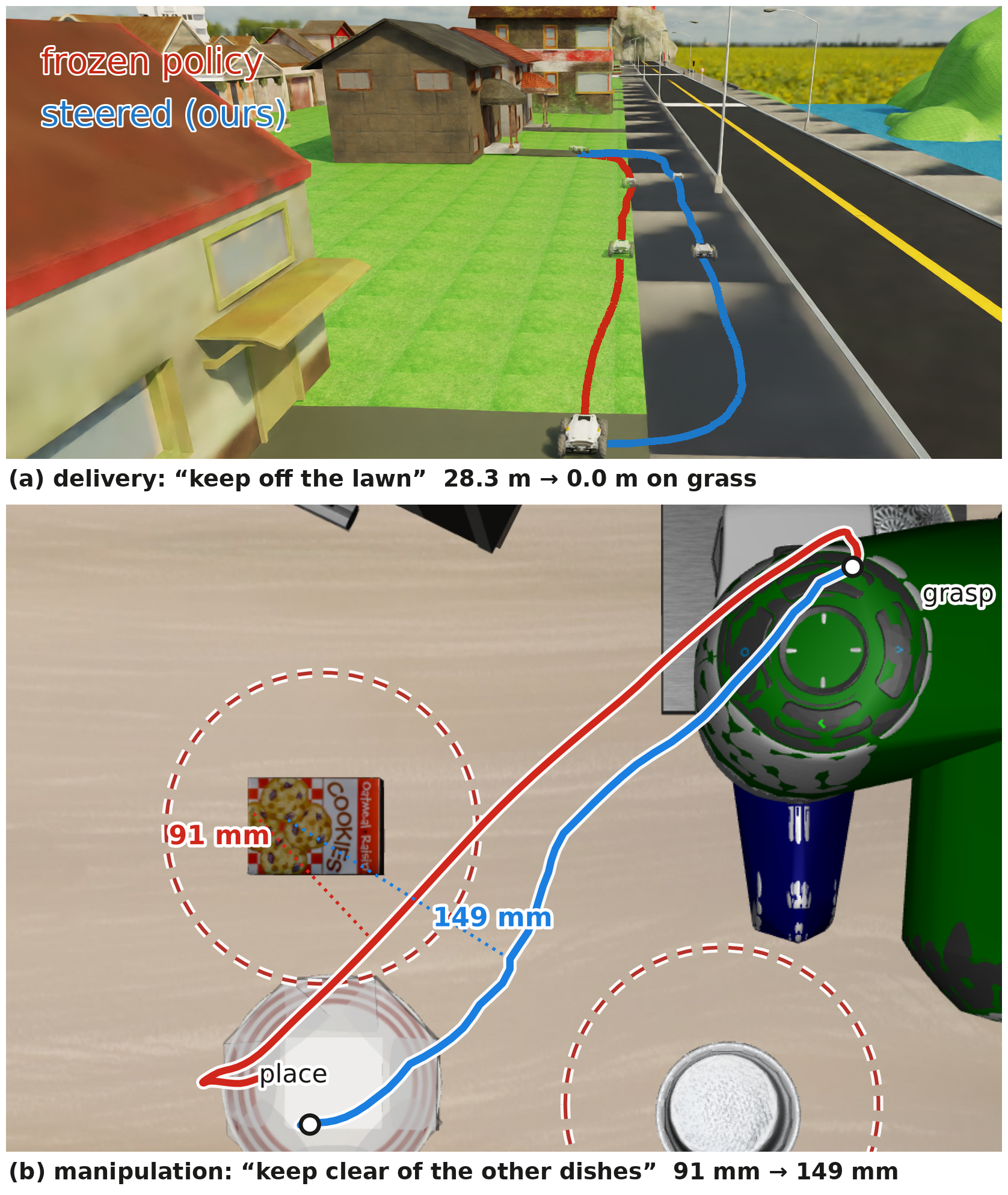}
\caption{Within each panel both rollouts come from the same policy at the
same initial state: red is that policy on its own, blue is the same policy with our
dynamic-barrier update applied to the sampler.}
\label{fig:teaser}
\end{figure}

Adapting a pretrained policy to these deployment-specific requirements raises two key challenges. First, adaptation should leave the pretrained policy \emph{frozen}: retraining or fine-tuning for every deployment requires additional data and computation and undermines the benefits of a broadly reusable pretrained policy. Instead, we adapt behavior at inference time by intervening in the sampling process of generative policies, steering their outputs toward behaviors that satisfy deployment-specific requirements and preferences while leaving the policy itself unchanged (\Cref{fig:teaser}).
Second, deployment objectives may be specified with a clear \emph{priority order} by the operator, user, or application, a structure not addressed by prior steering methods. For example, a delivery robot should stay off the lawn only when doing so does not require traversing a curb that its embodiment cannot climb. Here, the distinction between higher- and lower-priority objectives is structural: improvements in a lower-priority objective should never compensate for violating a higher-priority one. Weighted sums cannot guarantee this behavior because they assign objectives a finite exchange rate. Fixed weights can enforce a desired priority only when the possible ranges of the objectives are known, yet these ranges can vary substantially across deployment instances—for example, a longer delivery route may accumulate much more lawn cost than a shorter one. Consequently, weights that preserve the desired priority on one instance may fail to preserve it on another.

We address this problem by enforcing the priority order directly at inference time rather than assigning a fixed trade-off between objectives. Our method operates in two stages: it first steers individual samples and then selects which resulting sample to execute. 
Given a set of ordered runtime costs, our method applies dynamic-barrier guidance to steer each sampled trajectory while preserving higher-priority objectives to first order. It then applies a lexicographic selection cascade that filters the resulting candidates one priority level at a time.
Because both mechanisms operate on sampled trajectories rather than modifying the policy network, the pretrained policy remains frozen and the same approach applies across different iterative generative samplers.

To this end, we make the following contributions:
\begin{enumerate}
\item \textbf{Inference-time steering for lexicographically ordered objectives.}
We formulate steering frozen generative robot policies under lexicographically ordered runtime costs and develop a sampler-agnostic approach combining dynamic-barrier guidance with lexicographic candidate selection. The method applies to both diffusion and flow-matching policies without retraining or fine-tuning.

\item \textbf{Navigation under ordered feasibility and preference costs.}
We introduce a last-mile delivery benchmark that creates frequent conflicts between embodiment-specific traversability and user-specified surface preferences. Our method substantially improves success, preference compliance, and traversability over the frozen policy and achieves better compliance than tuned weighted-sum baselines.

\item \textbf{Cross-domain transfer and reduced tuning sensitivity.}
On LIBERO \cite{liu2023libero}, the same method transfers to a flow-matching manipulation policy and improves preference and feasibility compliance without sacrificing task success. When objective scales remain relatively consistent across episodes, as in this setting, a well-tuned fixed weight is expected to preserve the desired ordering, and our method accordingly matches its performance. A controlled D3IL \cite{jia2024d3il} study then shows the practical difference: the dynamic barrier maintains comparable best performance over a substantially wider range of parameter settings, while the effective fixed weight occupies a narrower range and changes with steering strength.
\end{enumerate}

\section{Related Work}
\subsection{Inference-Time Steering of Generative Policies}

Inference-time guidance modifies the behavior of a pretrained generative model without retraining its parameters. Gradient-based guidance was introduced for diffusion models~\cite{dhariwal2021diffusion} and later applied to trajectory generation~\cite{janner2022planning}. For frozen robot policies, existing methods steer generated trajectories through gradient updates~\cite{wang2025inference,zeng2025navidiffusor}, reweighting or resampling~\cite{singhal2025general,wu2023practical,liu2026vls}, or selection among generated candidates~\cite{olguin2026sem}. Other approaches explicitly impose safety constraints through barrier functions, projection, or constrained optimization~\cite{xiao2023safediffuser,zhang2026constrained,yin2023shield,li2026hardflow}. Recent work also considers inference-time modification of flow-matching policies~\cite{zhan2026seam}.

These approaches generally combine multiple objectives through a weighted cost or a single reward, or separate a single constraint from an otherwise scalar objective. In contrast, we consider multiple runtime costs with an explicit priority order. 

\subsection{Prioritized and Lexicographic Objectives}

Prioritized objectives have a long history in robotics and optimization, including admissibility-based local planning~\cite{fox1997dynamic}, task-priority control and hierarchical quadratic programming~\cite{slotine1991general,escande2014hierarchical}, rulebooks for autonomous driving~\cite{censi2019liability}, and control-barrier-function methods that prioritize safety over performance~\cite{ames2016control}. Lexicographic objectives have also been studied in reinforcement learning and optimization~\cite{gabor1998multi,wray2015multi,skalse2022lexicographic,zhang2023targeted}.

Our gradient update builds on dynamic-barrier optimization~\cite{gong2021automatic}, which modifies descent on one objective to enforce progress on a prioritized objective. Pareto Navigation Gradient Descent~\cite{ye2022pareto} extends a similar quadratic-program structure to several objectives, but treats them as an unordered Pareto set. We instead consider explicitly ordered runtime costs and combine prioritized gradient steering with a lexicographic candidate-selection rule inside the sampling process of a frozen generative policy.

\section{Method}
\subsection{Problem Setting}
\label{sec:problem_setting}

A frozen generative policy maps an observation $o$ in an observation space $\mathcal{O}$ to $S$
candidate action sequences through an iterative sampling process. We call a
single action sequence a \emph{sample} and write it as a point
$x\in\mathcal{X}=\mathbb{R}^{T\times d_a}$, a horizon of $T$ actions of
dimension $d_a$. Starting from noise $x^{(0)}\sim\mathcal{N}(0,I)$, each
candidate is refined by $K$ learned sampling steps
\begin{equation}
\label{eq:sampler}
x^{(k)}
=
\operatorname{Step}_{\theta}\!\left(x^{(k-1)},k,o\right),
\qquad
k=1,\dots,K,
\end{equation}
where $\operatorname{Step}_{\theta}:\mathcal{X}\times\{1,\dots,K\}\times\mathcal{O}\to\mathcal{X}$
is one step of the frozen sampler and may correspond, for example, to a
diffusion reverse step or a flow-matching integration step. We index the $S$ candidates by $i=1,\dots,S$, writing $x_i$ for the $i$-th sample, and we drop the
step superscript, writing $x$ for the current iterate when the step index is not
needed.

The runtime costs we wish to impose depend on where the
robot, or the object it carries, actually travels, rather than on the policy's
internal action encoding. We therefore evaluate them on a \emph{workspace
trajectory} $y\in\mathcal{Y}=\mathbb{R}^{T\times d_w}$, where $d_w$ is the
dimension of the workspace, and assume a differentiable decoder
\begin{equation}
\label{eq:decoder}
\Phi:\mathcal{X}\to\mathcal{Y},
\qquad
y=\Phi(x),
\end{equation}
that maps a sample to the trajectory it represents. 
In our navigation
experiments $d_w=2$ and $\Phi$ integrates the sampled waypoint displacements
into the world frame; in our manipulation experiments $d_w=3$ (LIBERO) or $d_w=2$ (D3IL) and $\Phi$ maps the sampled action chunk to the predicted path of the end effector.

At deployment, we introduce $L$ runtime costs
\begin{equation}
\label{eq:costs}
c_j:\mathcal{Y}\to\mathbb{R}_{\geq0},
\qquad
j=1,\dots,L,
\end{equation}
each continuously differentiable, together with a priority order
$c_1 \succ c_2 \succ \cdots \succ c_L$,
ordered from highest to lowest priority. The ordering is lexicographic:
improving a lower-priority objective must not come at the expense of a
higher-priority one. The costs may represent, for example, physical
feasibility, safety requirements, operator preferences, or other
deployment-specific objectives that were not available when the policy was
trained. Our experiments use two ordered costs, a feasibility cost
{$g:=c_1$} and a deployment preference {$f:=c_2$} with $g \succ f$, and
we present the method for this case; the construction extends to deeper
hierarchies, as discussed at the end of \Cref{sec:barrier_guidance}.

We keep the policy weights, sampling schedule, and observations fixed. Our
method requires only {the following three ingredients.}
{\begin{enumerate}
\item[(i)] Access to the sample $x\in\mathcal{X}$ between sampling steps.
\item[(ii)] The differentiable decoder $\Phi$ of \Cref{eq:decoder}, together
with a \emph{reconstruction map}
$
\Psi:\mathcal{Y}\times\mathcal{X}\to\mathcal{X}
$
that writes an edited trajectory back into a sample, overwriting only the channels that $\Phi$ reads and leaving the rest of the sample unchanged, so that
$
\Phi\!\left(\Psi(y,x)\right)=y
$
for all $x\in\mathcal{X}$ and all $y$ in the range of $\Phi$. We abbreviate
$\Psi(y,x)$ by $\Phi^{-1}(y)$ where the sample being overwritten is clear from
context.
\item[(iii)] Optionally, the policy's value head
$V:\mathcal{X}\times\mathcal{O}\to\mathbb{R}$, used only for final candidate
selection.
\end{enumerate}}

We impose the ordering at two stages, summarized in \Cref{alg:steering}. {\bf Lexicographic barrier guidance} modifies individual samples while preserving higher-priority objectives, and {\bf lexicographic candidate selection} applies the same ordering when choosing which candidate to execute.

\begin{algorithm}[t]
\caption{Lexicographically Steered Sampling (one planning step). {The parameterization $M$, step size $\eta$ and barrier constants $\alpha_b,\beta_b$ are defined in \Cref{sec:barrier_guidance}; the tolerances $\varepsilon_g,\varepsilon_f,\varepsilon_p$ and the progress score $p_i$ in \Cref{sec:lexicographic_selection}.}}
\label{alg:steering}
\small
\begin{algorithmic}[1]
\Require observation $o$; frozen sampler $\operatorname{Step}_{\theta}$; ordered costs $g \succ f$; decoder $\Phi$ and reconstruction $\Phi^{-1}$; parameterization $M$; step size $\eta$; barrier constants $\alpha_b,\beta_b$; tolerances $\varepsilon_g,\varepsilon_f,\varepsilon_p$
\State draw $S$ samples $x_1,\dots,x_S$ from noise
\For{$k = 1,\dots,K$}
\cmt{frozen sampler, unchanged}
  \State $x_i \gets \operatorname{Step}_{\theta}(x_i,k,o)$ for all $i$
\cmt{barrier guidance (\Cref{sec:barrier_guidance})}
  \For{each candidate $i$}
    \State $u_i \gets \arg\min_{u} \lVert M(u)-\Phi(x_i) \rVert^2$ \InlineComment{e.g.\ fit B-spline}
    \State $\phi \gets \min\!\big(\alpha_b\, g(u_i),\; \beta_b \lVert \nabla g(u_i) \rVert^2\big)$
    \State $\lambda \gets \max\!\big(\big(\phi - \langle \nabla f(u_i), \nabla g(u_i)\rangle\big) \big/ \lVert\nabla g(u_i)\rVert^2,\; 0\big)$
    \State $u_i \gets u_i - \eta\,(\nabla f(u_i) + \lambda\, \nabla g(u_i))$
    \State $x_i \gets \Phi^{-1}(M(u_i))$
  \EndFor
\EndFor
\cmt{guidance may run after every step or as repeated iterations on the final sample}
\cmt{selection (\Cref{sec:lexicographic_selection})}
\State $y_i \gets \Phi(x_i)$ 
\State $p_i \gets$ task progress of candidate $i$, for all $i$
\State $\mathcal{I} \gets \{1,\dots,S\}$, optionally dropping standstill candidates
\cmt{feasibility first}
\State $\mathcal{I} \gets \{\, i \in \mathcal{I} : g(y_i) \le \min_{i'\in\mathcal{I}} g(y_{i'}) + \varepsilon_g \,\}$
\cmt{then preference}
\State $\mathcal{I} \gets \{\, i \in \mathcal{I} : f(y_i) \le \min_{i'\in\mathcal{I}} f(y_{i'}) + \varepsilon_f \,\}$
\cmt{then task progress}
\State $\mathcal{I} \gets \{\, i \in \mathcal{I} : p_i \ge \max_{i'\in\mathcal{I}} p_{i'} - \varepsilon_p \,\}$
\State \Return the candidate in $\mathcal{I}$ preferred by the value head, if one exists
\end{algorithmic}
\end{algorithm}

\subsection{Lexicographic Barrier Guidance}
\label{sec:barrier_guidance}

\paragraph{Intervening between sampler steps}
We apply guidance directly to the sample returned by the frozen sampler,
\[
x \leftarrow \operatorname{Step}_{\theta}(x,k,o),
\qquad
x \leftarrow \operatorname{Steer}(x),
\]
{where $\operatorname{Steer}:\mathcal{X}\to\mathcal{X}$ is the update
derived below,} rather than modifying the network output used inside
$\operatorname{Step}_{\theta}$. The intervention therefore has the same
form for diffusion and flow-matching samplers.

\paragraph{{Editable coordinates}}
We edit a parameterization of the workspace trajectory: a vector of \emph{editable coordinates}
$u\in\mathcal{U}=\mathbb{R}^{m\times d_w}$ together with a differentiable
reconstruction map
\begin{equation}
\label{eq:param}
M:\mathcal{U}\to\mathcal{Y},
\qquad
y=M(u).
\end{equation}
For example, in our navigation task $u$ collects $m=7$ B-spline control points and $M$
is linear, $M(u)=Bu$ with $B\in\mathbb{R}^{T\times m}$ the clamped cubic
B-spline basis, so an update moves a few control points rather than every
waypoint individually. In our manipulation tasks the trajectory is edited
directly: $m=T$ and $M$ is the identity.

The runtime costs are defined on $\mathcal{Y}$, so we compose them with the
parameterization and, with a slight abuse of notation, write the composed costs
with the same symbols: $g(y)$ denotes the cost
on a workspace trajectory and $g(u)$ its composition with $M$,
\begin{equation}
\label{eq:pullback}
g(u):=c_1\!\left(M(u)\right),
\qquad
f(u):=c_2\!\left(M(u)\right).
\end{equation}
{Identifying $\mathcal{U}$ with $\mathbb{R}^{md_w}$ and $\mathcal{Y}$ with
$\mathbb{R}^{Td_w}$, the chain rule gives}
\begin{equation}
\label{eq:chain}
\nabla g(u)
=
J_M(u)^{\!\top}\,\nabla c_1\!\left(M(u)\right),
\end{equation}
where $J_M(u)$ is the Jacobian of $M$ at $u$, equal to the constant matrix
$B$ in the linear case, and likewise for $f$. All gradients below are taken with
respect to $u$, and $\langle\cdot,\cdot\rangle$ and $\lVert\cdot\rVert_2$ are the
Euclidean inner product and norm on $\mathcal{U}$. After editing $u$, we
reconstruct the sample as
$
x \leftarrow \Phi^{-1}\!\left(M(u)\right).
$

\paragraph{Dynamic barrier condition}
{Fix $u\in\mathcal{U}$, let $d\in\mathcal{U}$ be an update direction and
$\eta>0$ a step size, and write $u^{+}=u-\eta d$. Since $g$ is continuously
differentiable, the first-order change in the prioritized cost is}
\begin{equation}
\label{eq:first_order}
g(u^{+})-g(u)
=
-\eta
\left\langle
\nabla g(u),d
\right\rangle
+o(\eta).
\end{equation}
Consequently, the condition
$\langle \nabla g(u), d \rangle \geq 0$
prevents $g$ from increasing, up to first order. A dynamic barrier
strengthens this condition by requiring a positive decrease whenever
the prioritized cost is away from its desired value.

Following dynamic-barrier gradient descent~\cite{gong2021automatic}, we
define the {dynamic barrier function
$\phi:\mathcal{U}\to\mathbb{R}_{\geq0}$ by}
\begin{equation}
\label{eq:phi}
\phi(u)
:=
\min\left(
\alpha_b\, g(u),\;
\beta_b
\left\|
\nabla g(u)
\right\|_2^2
\right),
\qquad
\alpha_b,\beta_b>0.
\end{equation}
Gong et al.\ write the first term as
$\alpha_b(g(u)-\widehat g)$, where $\widehat g$ is a known lower
bound {on $g$}. Our runtime costs are non-negative and use zero as their
desired value, so we set $\widehat g=0$. {Since $g(u)\geq0$ and
$\alpha_b,\beta_b>0$, we have $\phi(u)\geq0$ for every $u\in\mathcal{U}$,
with $\phi(u)=0$ exactly when $g(u)=0$ or $\nabla g(u)=0$.}

The two terms determine how much decrease we require from the prioritized cost.
The first term, $\alpha_b\, g(u)$, requires a larger decrease when the current
cost is large. The second term, $\beta_b\|\nabla g(u)\|_2^2$, limits this
requirement when the local gradient is small. Taking the minimum prevents the
barrier from demanding more progress than the local cost gradient can support.

The corresponding barrier condition is
$\langle \nabla g(u), d \rangle \geq \phi(u)$.
{Substituting it into \Cref{eq:first_order} gives}
\begin{equation}
\label{eq:decrease}
g(u^{+})-g(u)
\leq
-\eta\,\phi(u)+o(\eta)
\leq
o(\eta),
\end{equation}
{so any direction satisfying the barrier condition decreases the
prioritized cost to first order, strictly so whenever $\phi(u)>0$.}

\paragraph{Steering update}
The steering direction is the one closest to ordinary descent on the
preference while satisfying the feasibility barrier:
\begin{equation}
\label{eq:barrier_qp}
d^{\star}(u)
=
\arg\min_{{d\in\mathcal{U}}}
\frac{1}{2}
\left\|
d-\nabla f(u)
\right\|_2^2
\quad
\text{s.t.}
~
\left\langle
\nabla g(u),d
\right\rangle
\geq
\phi(u).
\end{equation}
When $\nabla g(u)\neq0$, {the feasible set is a non-empty half-space and
this one-constraint quadratic program has the} closed-form solution
\begin{equation}
\label{eq:barrier_update}
d^{\star}(u)
=
\nabla f(u)
+
\lambda(u)\,\nabla g(u),
\end{equation}
{with multiplier}
\begin{equation}
\label{eq:lambda}
\lambda(u)
=
\max\left(
\frac{
\phi(u)-
\left\langle
\nabla f(u),\nabla g(u)
\right\rangle
}{
\left\|
\nabla g(u)
\right\|_2^2
},
\;0
\right),
\end{equation}
and the update is
$
u
\leftarrow
u-\eta\, d^{\star}(u).
$
If
$\langle \nabla f(u), \nabla g(u) \rangle \geq \phi(u)$,
ordinary descent on $f$ already satisfies the feasibility barrier, so
$\lambda(u)=0$. Otherwise, $\lambda(u)>0$ adds the smallest correction along
$\nabla g(u)$ needed to make the barrier active,
$\langle \nabla g(u), d^{\star}(u) \rangle = \phi(u)$.
If $\nabla g(u)=0$, then $\phi(u)=0$ and the barrier contains no local
information with which to alter the trajectory. We therefore set
$\lambda(u)=0$ and recover ordinary descent on $f$. In implementation, we
apply the same branch when
$\|\nabla g(u)\|_2^2$ falls below a small numerical threshold, rather
than adding that threshold to the theoretical denominator.

The multiplier is therefore dynamic rather than fixed: it depends on
the current value of the feasibility cost, the magnitude of its
gradient, and its local alignment with the preference gradient. It
vanishes when preference descent already respects feasibility and
increases only when required to enforce the higher-priority barrier.

\paragraph{Generalization to deeper hierarchies}
The construction above extends to an ordered set of costs
$
c_1 \succ c_2 \succ \cdots \succ c_L,
$
where $c_L$ is the objective being optimized and each higher-priority
cost $c_j$, $j<L$, has its own barrier requirement
{$\phi_j(u):=\min(\alpha_j c_j(u),\beta_j\|\nabla c_j(u)\|_2^2)$}. If all
barrier conditions are compatible, we can enforce them simultaneously:
\[
\left\langle \nabla c_j(u),d \right\rangle
\geq \phi_j(u),
\qquad j=1,\ldots,L-1,
\]
while choosing the direction closest to ordinary descent on $c_L$.
A similar multi-constraint QP appears in Pareto Navigation Gradient
Descent~\cite{ye2022pareto}, where the constrained objectives are
treated symmetrically.

When the barrier conditions conflict, {that is, when the intersection of the
half-spaces above is empty,} enforcing them simultaneously
does not specify which condition should be relaxed. To preserve the
priority order, a non-negative slack $\xi_j$ can be introduced at each level,
\[
\left\langle \nabla c_j(u),d \right\rangle
\geq \phi_j(u)-\xi_j,
\qquad \xi_j\geq0,
\]
and the slacks can be minimized lexicographically: first $\xi_1$, then $\xi_2$
without worsening the optimum at level $1$, and so on. Only after the
best achievable slack has been fixed at every prioritized level do we
choose the direction closest to $\nabla c_L{(u)}$. Thus, when the barrier
conditions conflict, a lower-priority condition is relaxed only after
all higher-priority conditions have been satisfied as well as possible.

The two-cost case used in our experiments is the special case $L=2$.
There is then only one prioritized cost, $c_1=g$, and one objective to
optimize, $c_2=f$. Its single barrier condition is {feasible whenever
$\nabla g(u)\neq0$}, so the optimal slack is zero and the
general formulation reduces exactly to the one-constraint QP in
\Cref{eq:barrier_qp}, with the closed-form update in
\Cref{eq:barrier_update}.

\subsection{Lexicographic Candidate Selection}
\label{sec:lexicographic_selection}

Barrier guidance modifies each candidate trajectory independently. The policy still
produces $S$ candidates, so we apply the same priority ordering again when choosing which
candidate to execute. This selection stage can include several lexicographic levels. For
example, in navigation we consider feasibility, preference, task progress, and
finally the policy's value head.

{Let $y_i=\Phi(x_i)$ denote the workspace trajectory of candidate $i$ and let
$\mathcal{I}_1\subseteq\{1,\dots,S\}$ denote the initial set of candidate
\emph{indices}.} We filter this set sequentially according to the ordered costs
$
c_1 \succ c_2 \succ \cdots \succ c_L.
$
For each priority level $j$, let
\[
c_j^{\min}
=
\min_{i\in\mathcal{I}_j}
c_j(y_i)
\]
be the best cost among the candidates that survived the previous levels. We retain all candidates whose cost is within a tolerance $\varepsilon_j{>0}$ of this minimum:
\begin{equation}
\label{eq:filter}
\mathcal{I}_{j+1}
=
\left\{
i\in\mathcal{I}_j
\;:\;
c_j(y_i)
\leq
c_j^{\min}+\varepsilon_j
\right\}.
\end{equation}
The tolerance $\varepsilon_j$ is defined in the units of cost $c_j$. It prevents small differences in noisy runtime costs from causing brittle decisions while preserving the priority ordering: a candidate rejected at level $j$ cannot be recovered by performing better on any lower-priority objective.

Because the threshold is defined relative to the minimum over the current candidate set, {the minimizer at level $j$ always survives, so
$\emptyset\neq\mathcal{I}_{j+1}\subseteq\mathcal{I}_j$ and} at least one candidate always survives each stage.


After all prioritized costs have been considered, we rank the remaining candidates by task progress. Let $p_i{\in\mathbb{R}}$ denote the progress score of candidate $i$, with larger values indicating better progress. We retain candidates satisfying
\[
p_i
\geq
p^{\max}-\varepsilon_p,
\qquad
p^{\max}
=
\max_{i\in\mathcal{I}_{L+1}} p_i.
\]
If the frozen policy provides a value head, we use it only as the final tie-break among the remaining candidates.

The resulting cascade is lexicographic because each objective is evaluated only over candidates admitted by all higher-priority levels. Lower-priority objectives can therefore distinguish between candidates that are approximately equivalent at higher-priority levels, but they cannot compensate for a worse higher-priority cost.

Barrier guidance and candidate selection impose the same ordering at complementary stages of inference. The barrier modifies each candidate while respecting prioritized costs, and the selection cascade prevents the final choice from reintroducing a trade between those costs.

\section{Last-Mile Delivery with a Diffusion Policy}
\begin{figure}[t]
\centering
\includegraphics[width=0.49\textwidth]{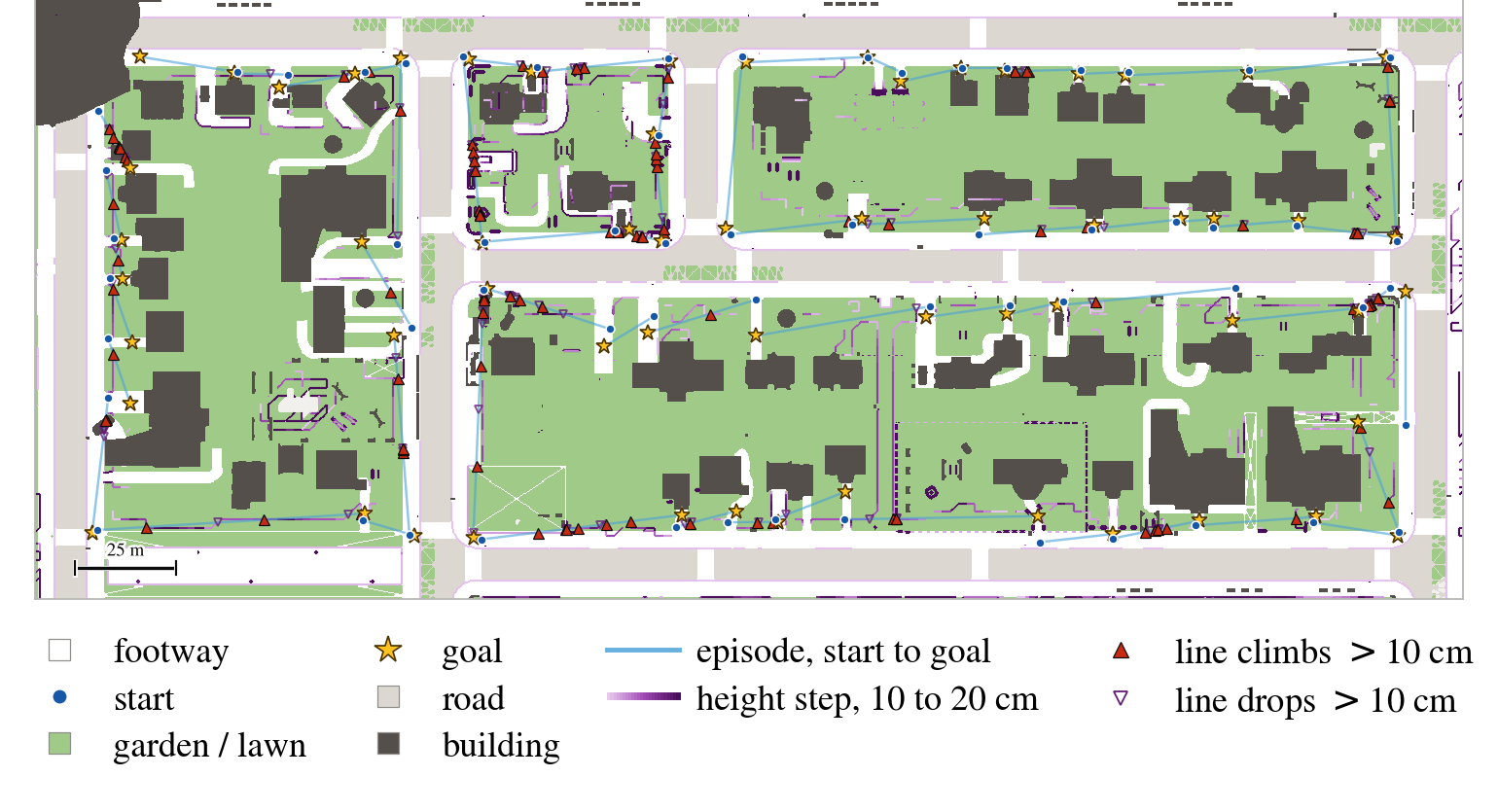}
\caption{The navigation benchmark: 60 door-to-door episodes in CARLA Town01, rebuilt in
Isaac Sim.}
\label{fig:overhead}
\end{figure}

We build a navigation benchmark inspired by last-mile delivery, where a robot is tasked with reaching the front doors of different residences. The robot must account for both what its embodiment can traverse and where it is permitted or preferred to drive. For example, a residential lawn is physically traversable but may be undesirable, whereas a tall curb may be physically impassable for the deployed robot. These two considerations therefore have different priorities: the robot should avoid the lawn when possible, but should never do so at the expense of a route it cannot traverse. We construct episodes in which the direct route often encounters both types of constraints (see \Cref{fig:overhead}).

\begin{figure*}[t]
\centering
\includegraphics[width=0.95\textwidth]{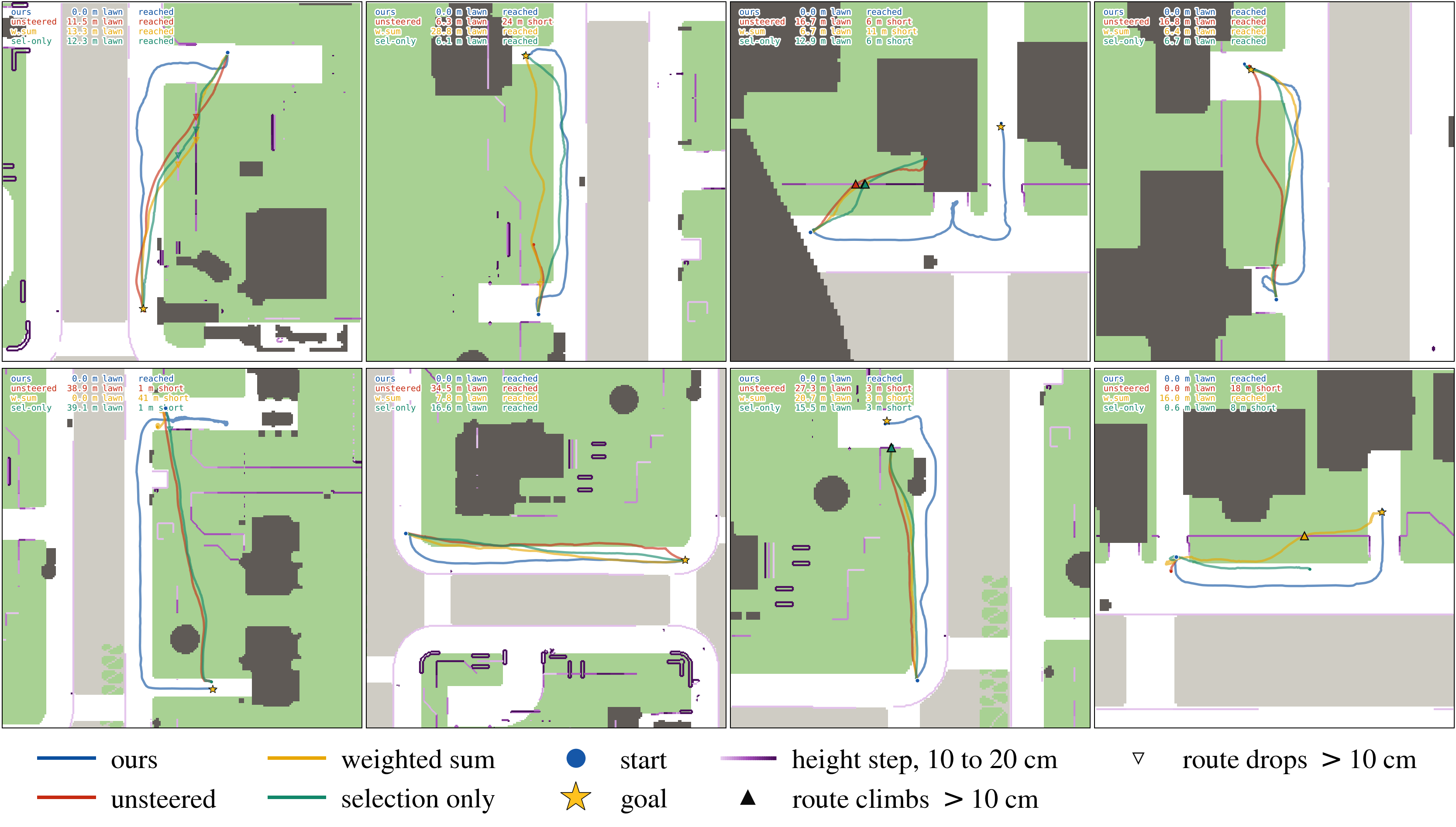}
\caption{Selected episodes of the delivery task. Labels give meters driven on lawn and whether the goal was
reached.  Ours drives $0.0$\si{m} of lawn and reaches the goal in all eight; the
baselines drive up to $39$\si{m} of lawn, and the two that cross no lawn do so by ending $41$ and
$18$\si{m} short.  Purple marks a $10$--$20$\si{cm} height step on drivable ground, shaded by
magnitude; triangles mark where a route climbs ($\blacktriangle$) or drops
($\triangledown$) more than $10$\si{cm}.  Note that every triangle sits on a baseline. Ours sometimes drives further than needed (row 1 column 3 and row 2 column 1): policy and steering are both local, so a compliant shortcut can only be ruled out by probing for it.  }
\label{fig:navigation_showcase}
\end{figure*}

\subsection{Simulation Setup}
\label{sec:benchmark}
\emph{Simulator and Scene.} We evaluate in NVIDIA Isaac Sim \cite{isaacsim} on a residential town we took from \cite{roth2024viplanner}, which comes from CARLA Town01 \cite{dosovitskiy2017carla}.

\emph{Robot.} We use an AgileX Scout 2.0 in the simulation, which is equipped with an RGB-D camera and a LiDAR. Our step-climbing tests indicate that \SI{0.10}{\m} is a conservative recommended upper bound on traversable step height.

\emph{Episodes.} We select $60$ door-to-door start--goal pairs spanning $9.2$--\SI{66}{\m} (median \SI{23.4}{\m}) along the sidewalks (see \Cref{fig:overhead}). To ensure a fair test of both deployment requirements, we verify that every pair admits a feasible pavement-only route wide enough for the robot; the resulting route is at most $1.61\times$ the corresponding free-space route (median $1.04\times$). At the same time, we deliberately select pairs whose direct shortcuts expose the two constraints: $55/60$ straight-line routes cross a lawn, while $43/60$ encounter a step $\geq$ \SI{0.10}{\m}. 
Thus, the benchmark provides feasible alternatives while systematically exposing the policy to conflicts between the two requirements.

\emph{Metrics.} An episode is considered successful if the robot comes within \SI{1}{\m} of the goal within \SI{300}{\s}. We report the following metrics:
(i) \emph{success}, the number of successful episodes out of 60;
(ii) \emph{on-lawn distance}, the mean lawn distance per episode measured at the robot center;
(iii) \emph{zero-lawn episodes}, with less than \SI{0.05}{\m} of lawn contact;
(iv) \emph{traversability failures}, failed episodes whose stall point contains a $\geq$ \SI{0.10}{\m} step under the robot footprint;
(v) \emph{curb episodes}, episodes with any travel within one half-width of a road cell; and
(vi) \emph{driven distance}.

\subsection{Implementation}
\label{sec:nav_impl}
We use NavDP~\cite{cai2025navdp} as our base policy. NavDP is a goal-conditioned RGB-D diffusion policy with a learned value head. It samples $S{=}16$ candidate trajectories, each consisting of $T{=}24$ waypoints, over $K{=}10$ DDPM steps, and executes the selected trajectory using a model-predictive controller. 

\emph{Trajectory Representation.}
We represent each sampled trajectory as a position sequence $y\in\mathbb{R}^{T\times 2}$ by cumulatively
summing its waypoint displacements and transforming it from
NavDP’s camera frame to the world frame. For gradient
refinement, we fit this path using a clamped cubic B-spline
basis and optimize the spline control points. The refined path
is then mapped back to waypoint displacements, while the
sample’s yaw channel is left unchanged. 

\emph{Cost Fields.}
The preference cost $f$ is obtained from a semantic class map. We use a \SI{0.35}{\m} ramp around the lawn boundary, corresponding to the robot's half-width, so that cost begins when the footprint reaches the lawn; bilinear interpolation makes the field differentiable. The feasibility cost $g$ is constructed online from lidar as a traversability risk field, following the approach of STEP~\cite{fan2021}. It marks steps at or above the \SI{0.10}{\m} bound as high risk and assigns unknown-risk to occluded cells, where negative obstacles may be hidden. We aggregate risk using conditional value-at-risk (CVaR)~\cite{rockafellar2000optimization}, taking the mean of the worst $20\%$ of values within a \SI{0.4}{\m} footprint disc.

\emph{Steering.}
Although the barrier update of \Cref{eq:barrier_update} can be applied between any two sampler steps, we apply all barrier updates after the final DDPM step. Applying them earlier allows subsequent denoising steps to partially undo the changes. We refine each completed sample for 15 barrier-update iterations in a clamped cubic B-spline basis with $7$ control points. Each control-point update is limited by a \SI{0.15}{\m} trust region, and we use $\eta = 1.0$ and $\alpha_b = \beta_b = 1$. The first control point is pinned to the robot, and each update is projected perpendicular to the local path tangent to prevent the refinement from reducing cost simply by retreating along the path.


\emph{Selection.}
We select the executed trajectory using the lexicographic cascade with a horizon discount $\gamma_h = 0.85$ on the feasibility cost, feasibility tolerance $\varepsilon_g = 0.1$, preference tolerance $\varepsilon_f = 0.05$, and progress tolerance $\varepsilon_p = 1.5$\si{\m}. The policy's value head is used only at the final stage. We discount the feasibility cost so that unknown-risk in the far tail does not make all candidates indistinguishable.

\emph{Goal Conditioning.}
NavDP's point-goal representation covers $[0,10]\times[-10,10]$\si{\m} in forward and lateral coordinates, whereas our delivery goals can be up to \SI{66}{\m} away. Rather than independently clipping the two coordinates, which can substantially distort the goal bearing, we rescale goals farther than \SI{10}{\m} down to \SI{10}{\m} while preserving their bearing; nearer goals pass through unchanged. The original, unclipped goal is retained separately for the cascade's progress stage. We apply this re-encoding to all experimental arms, including the unsteered baseline, so it is not part of the steering method. After the correction, every goal ahead of the robot is encoded with its true bearing. One case remains unrepresentable: a goal \emph{behind} the robot has a negative forward coordinate, which the policy's encoder still clips into its forward range. This residual defect belongs to the frozen policy's interface, not to the steering, and it produces a specific failure mode that we quantify in \Cref{sec:failure_analysis}.

\begin{table*}[t]
\caption{Navigation benchmark.\\ 60 episodes, three repeats per method (mean\,$\pm$\,sd).}
\label{tab:headline}
\centering
\small
\begin{tabular}{lcccccc}
\toprule
Method & Success $\uparrow$ & Lawn (m) $\downarrow$ & Zero-lawn $\uparrow$ & Trav.\ failures $\downarrow$ & Curb ep. $\downarrow$ & Driven (m) \\
\midrule
Unsteered & $44.3\pm5.0$ & $9.24\pm0.61$ & $17.3\pm0.6$ & $12.0\pm4.6$ & $3.3\pm0.6$ & ${27.2\pm0.8}$ \\
\midrule
Selection Only & $48.0\pm2.6$ & $7.01\pm0.42$ & $23.3\pm1.5$ & $9.0\pm3.0$ & $0.3\pm0.6$ & $27.2\pm1.4$ \\
\midrule
\multicolumn{7}{l}{\emph{Weighted Sum}}\\
\quad $(w_p, \eta) = (100, 0.10)$ & $51.0\pm1.0$ & $6.58\pm0.34$ & $22.7\pm1.5$ & $6.0\pm0.0$ & $\mathbf{0.0\pm0.0}$ & $27.5\pm1.0$ \\
\quad $(w_p, \eta) = (50, 0.10)$ & $49.7\pm1.5$ & $6.84\pm0.40$ & $22.3\pm1.2$ & $8.3\pm2.3$ & $\mathbf{0.0\pm0.0}$ & $27.4\pm1.0$ \\
\quad $(w_p, \eta) = (25, 0.20)$ & $50.7\pm3.1$ & $6.34\pm0.14$ & $23.0\pm1.7$ & $5.7\pm2.5$ & $0.3\pm0.6$ & $27.7\pm0.9$ \\
\midrule
Ours & $\mathbf{56.0\pm1.0}$ & $\mathbf{2.60\pm0.24}$ & $\mathbf{49.3\pm2.1}$ & $\mathbf{2.0\pm1.0}$ & $\mathbf{0.0\pm0.0}$ & ${36.9\pm2.9}$ \\
\bottomrule
\end{tabular}
\vspace{2pt}
\end{table*}

\subsection{Baselines}
We compare against three baselines, and run every reported configuration for three repeats over the same 60 episodes.

\emph{Unsteered}: the frozen NavDP policy, unmodified except for the goal re-encoding described above.
 
\emph{Weighted Sum}: steers with $\nabla(w_t\,g + w_p\,f)$. Only the ratio of the two weights changes the direction of the update, so we fix
$w_t{=}1$ and search $w_p$ and the step size $\eta$ over a 12-point grid, on the
seven episodes our proposed method was tuned on.
We keep only the settings that lose no episode against the unsteered policy, and rank what remains by mean lawn.
We then choose the top three settings to run on the benchmark: $(w_p, \eta) \in \{(100, 0.10),\, (50, 0.10),\, (25, 0.20)\}$.

\emph{Selection Only}: our lexicographic cascade applied to NavDP's original candidates, with gradient steering disabled. The candidates are left unchanged, so this baseline isolates the effect of ordered candidate selection from the trajectory modification introduced by our full method.

\subsection{Results}
\subsubsection{Baseline comparisons}
Table~\ref{tab:headline} compares our method with the unsteered policy and weighted-sum guidance. Steering substantially improves preference compliance: mean on-lawn distance decreases from $9.24$ to $2.60$\si{\m} per episode, while zero-lawn episodes increase from $17.3$ to $49.3$ out of $60$. 
\Cref{fig:navigation_showcase} shows representative episodes.

Steering also improves feasibility. Across three repeats, the frozen policy fails $47$ episodes ($15.7$ per repeat), of which $36$ ($12.0$ per repeat) stall at steps of at least \SI{0.10}{\m} under the robot footprint, compared with only $2.0$ per repeat after steering (see \Cref{sec:failure_analysis} for failure analysis).

Weighted-sum guidance closes much of the success gap, achieving $49.7$--$51.0$ successful episodes compared with $56.0$ for our method. However, the lexicographic ordering gives substantially better compliance and feasibility: mean on-lawn distance is $2.60$\si{\m} versus $6.34$--$6.84$\si{\m}, zero-lawn episodes are $49.3$ versus $22.3$--$23.0$, and traversability failures are $2.0$ versus $5.7$--$8.3$ per repeat. The main cost of enforcing this ordering is path length, with our method driving approximately $9$\si{\m} further per episode.


Selection alone provides a smaller but consistent improvement. Applying the lexicographic cascade to NavDP's original candidate set, without gradient steering, raises success from $44.3$ to $48.0$ episodes, reduces mean on-lawn distance from $9.24$ to $7.01$\si{\m}, and lowers traversability failures from $12.0$ to $9.0$ per repeat. However, driven distance remains unchanged at $27.2$\si{\m}, indicating that candidate selection can only choose among behaviors already proposed by the frozen policy. The larger gains of the full method therefore come from the gradient update creating new, more compliant trajectories rather than selection alone.

\subsubsection{Failure analysis}
\label{sec:failure_analysis}

Twelve of our $180$ episodes fail. Three are simulator artifacts: the vehicle becomes stuck at its spawn because of interactions between the simulator physics and scene geometry. 
Six failures are wanders that occur when the goal lies \emph{behind} the robot. This is the one case the goal re-encoding of \Cref{sec:nav_impl} cannot repair: the encoder clips the negative forward coordinate into its forward range, so the conditioning no longer points toward the true goal and the robot loses a consistent direction.
The remaining three failures expose a limitation of local steering without global guidance. In these episodes, steering moves the robot into a locally low-cost region, but that region later becomes a dead end: reaching the goal requires returning to the pavement, which the robot cannot climb onto from its current position. The runtime costs therefore favor locally compliant motion without representing the downstream reachability of the resulting state. Avoiding these traps requires longer-horizon or global guidance in addition to the local inference-time steering used here.

\section{Manipulation with a Flow-Matching Policy}
Our second set of experiments asks two questions. First, does the same inference-time
steering mechanism transfer to a different generative process and action space? Second, when does the priority ordering actually matter?

The navigation benchmark rewards the ordering because the size of the two costs changes a
lot between episodes. A long delivery route can accumulate far more lawn cost than a short
one, so a weight that keeps the order on one episode may not keep it on another. 
By contrast, the pretrained manipulation policies and benchmarks we evaluate operate over
short-horizon tasks, where motion distances and cost magnitudes remain within a more
limited range. In this regime, a single well-tuned weight can more easily preserve the desired priority
across episodes, so we expect weighted-sum steering to match our method at its best setting.

We test this expectation directly. We first show that the proposed method transfers to a
flow-matching manipulation policy on LIBERO, where it performs on par with a tuned weighted
sum. We then run a controlled study on a fixed obstacle field, where we can sweep both
methods over their own parameters and measure how much tuning each one needs. The result is
that the two methods reach the same best performance, but the weighted sum only reaches it
inside a narrow band of weights, while our method reaches it over a wide range of settings.

\subsection{Transfer to a Flow-Matching Policy}

\emph{Policy and Environment.}
We evaluate the LIBERO-finetuned $\pi_0$ policy~\cite{black2024pi0}. The policy generates a
$T=50$-step action chunk of 6-DoF end-effector pose deltas and gripper commands by
integrating a learned flow field for $K{=}10$ Euler steps; the controller executes the
first $10$ actions before replanning. We sample $S{=}16$ candidates at every replan.

\emph{Trajectory Representation.}
Our costs are defined on the carried object's path rather than directly on the action
chunk. A naive integration of the commanded pose deltas substantially overestimates the
executed motion because the operational-space controller realizes only a fraction of each
command ($315$\si{\mm} predicted versus $98$\si{\mm} executed at the chunk end,
median over $102$ recorded chunks). 
We therefore fit a per-axis execution gain $\rho=(0.208,\,0.253,\,0.209)$, which reduces the
median terminal prediction error to $19$\si{\mm}. We incorporate this gain into the
sample-to-workspace map $\Phi$ and discount costs along the predicted horizon with
$\gamma_h=0.8$. We additionally inflate the cost margins by the $90$th-percentile surrogate
residual. This calibration is important: without it, steering optimizes a predicted path
substantially longer than the one executed by the robot.

\emph{Costs.}
Both runtime costs are defined on the carried object. The preference cost $f$ penalizes
carrying the object close to the avoid-set obtained from the task specification. The cost is smoothly attenuated near the placement
target so that it does not oppose the intended placement. The feasibility cost $g$
penalizes penetration of the table plane and other solid objects, excluding the placement
target. 

\emph{Steering and Selection.}
We apply the barrier update after each
Euler step in position space, using a \SI{0.004}{\m} trust region, $\eta=0.05$, $\alpha_b=10$, and $\beta_b=1$. The trust region is necessary to keep the edited samples within the
controller's executable range. Because $\pi_0$ has no value head, candidate selection
consists only of the feasibility and preference stages of the lexicographic cascade.

\emph{Baselines and Metrics.}
We evaluate on LIBERO using $80$ paired initial states across eight tasks, repeating each steered configuration three times under different sampling noise ($320$ episodes per repeat over the four methods); \Cref{tab:libero} reports mean$\pm$std over repeats.
We compare against \texttt{Unsteered} (the original $\pi_0$ policy), \texttt{Selection Only}
(the same $16$ candidates ranked by our cascade with gradient steering disabled), and
\texttt{Weighted Sum} (steering with $\nabla f + w_g \nabla g$ in place of the barrier's
state-dependent multiplier, under the same trust region and gain, with $w_g{=}100$ chosen
by a sweep on held-out calibration tasks and its candidates ranked by the same cascade).

The eight evaluation tasks are chosen before any steered run using only frozen-policy
rollouts. We keep tasks in which the preference is active during carrying and in which the
two cost gradients interact over a nontrivial part of the trajectory. We also require the
preference cost to vanish at both the start and the placement target, so that unavoidable
initial or terminal contact is not counted as a violation. All steering parameters are
calibrated on two held-out tasks excluded from evaluation.

We report success; preference exposure $E$, defined as the carried-object path length
inside the keep-out region after grasp; peak feasibility violation $g_{\max}$; and regrasp rate. 

\begin{table}[t]
\caption{LIBERO benchmark.}
\label{tab:libero}
\centering
\scriptsize
\setlength{\tabcolsep}{4pt}
\begin{tabular}{lccccc}
\toprule
Method & Success $\uparrow$ & $E$ (m) $\downarrow$ & $g_{\max}$ $\downarrow$ & Regrasp $\downarrow$ \\
\midrule
Unsteered   & $0.72 \pm 0.02$ & $0.408 \pm 0.005$ & $0.213 \pm 0.192$ & $0.08 \pm 0.03$ \\
S. Only & $0.79 \pm 0.01$ & $0.401 \pm 0.010$ & $0.119 \pm 0.097$ & $0.07 \pm 0.02$ \\
W. Sum    & $0.78 \pm 0.03$ & $\mathbf{0.348 \pm 0.009}$ & $0.068 \pm 0.035$ & $0.11 \pm 0.04$ \\
Ours            & $\mathbf{0.81 \pm 0.02}$ & $0.355 \pm 0.007$ & $\mathbf{0.066 \pm 0.029}$ & $\mathbf{0.07 \pm 0.01}$ \\
\bottomrule
\end{tabular}
\tabnote{$E$ is meters of preference violation; $g_{\max}$ is the max feasibility violation. Regrasp counts grasps after the first, i.e. how often the grip was lost mid-task.}
\end{table}

\emph{Results.}
Relative to the frozen policy, our method improves preference and feasibility at the same
time (Table~\ref{tab:libero}). Preference exposure decreases from $0.408$ to
$0.355$\si{\m}, and peak feasibility violation from $0.213$ to $0.066$. 
Success increases from $0.72$ to $0.81$, so the compliance
gains do not come at the cost of task completion.

\texttt{Selection Only} helps feasibility and success but leaves preference exposure
unchanged ($0.408$ versus $0.401$\si{\m}). The preference improvement therefore comes from
modifying the sampled trajectories, not from choosing among more candidates. This matches
navigation: selection contributes, but gradient steering is what produces behavior outside
the frozen policy's original candidate set.

\texttt{Weighted Sum} performs on par with our method here. The two are indistinguishable
in success, preference exposure, and peak feasibility violation, and we do not claim
superiority over a tuned weighted sum in this domain. This is the outcome we expect when
the cost scales do not vary much across episodes. What the transfer result establishes is
that the same method runs unchanged on a flow-matching policy and reaches the performance
of a separately tuned fixed weight without a task-specific weight sweep.

\begin{figure*}[t]
\centering
\includegraphics[width=\textwidth]{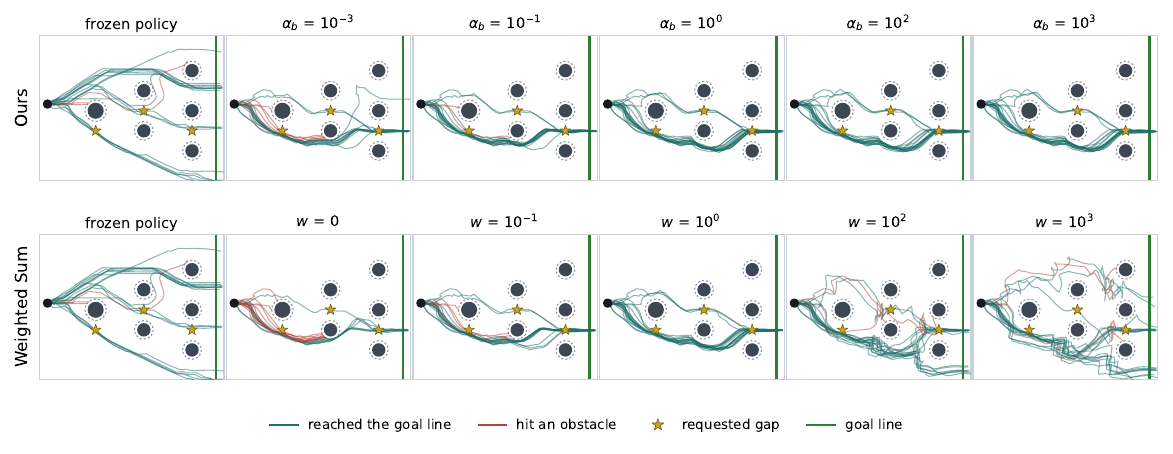}
\caption{Executed paths on D3IL \texttt{avoiding} as each method's parameter is varied
over
a wide range of values, at the same step size. Top-down view; the robot travels left to
right from the black dot to the goal line, and the stars mark the requested gap in each
obstacle row.
Our panels look nearly the same from $\alpha_b{=}10^{0}$ onward, while the weighted
sum only produces the requested behavior near $w{=}1$: at $w{=}0$ it does not enforce
feasibility, and at large $w$ the feasibility term dominates, the paths detour around
the obstacles, and compliance falls back to roughly the level it had with no feasibility
term at all. Paths are from one of the four requested routes.}
\label{fig:d3il_traj}
\end{figure*}

\subsection{How Much Tuning Does Each Method Need?}
\label{sec:sensitivity}
The LIBERO comparison shows parity at one operating point, but it does not show how hard
that point was to find. We therefore run a controlled study on a task where we can sweep
both methods over their own parameters and repeat every setting.

\emph{Task and Policy.}
We use the \texttt{avoiding} task from D3IL~\cite{jia2024d3il}. A robot arm must move a rod
from a start position to a goal line while passing through a field of six fixed cylindrical
obstacles arranged in three rows (see \Cref{fig:d3il_traj}). Contact with any obstacle ends the episode as a failure. The gaps between obstacles form $24$ distinct
routes, and the demonstrations use all of them. Our policy is an encoder--decoder diffusion
policy trained on the $96$ human demonstrations, producing $T=4$-step chunks of end-effector
positions over $K=16$ denoising steps. We train the policy with three seeds and pick the one 
with the highest success rate ($92.7\%$).

This task has the property we want to study. All six obstacles have almost the same radius
and the gaps between them are the same width, so the feasibility
cost has one characteristic scale across the whole workspace.

\emph{Costs.}
The feasibility cost $g$ is the simulator's own collision test made differentiable: a log
barrier on the distance between the rod surface and each obstacle surface, active inside a
\SI{0.02}{\m} band. The preference cost $f$ asks the policy to take a nominated route,
penalizing lateral distance from the chosen gap in each of the three rows. Because both
costs are anchored to the same obstacles, moving sideways to satisfy the preference moves
the rod toward an obstacle, so the two gradients oppose each other directly.

\emph{Metrics.}
We report \emph{success} (reaching the goal line without contact) and \emph{gate
agreement}, the fraction of the three rows where the robot passes through the requested
gap. Gate agreement measures how well the preference is satisfied. 

\begin{figure}[t]
\centering
\includegraphics[width=0.35\textwidth]{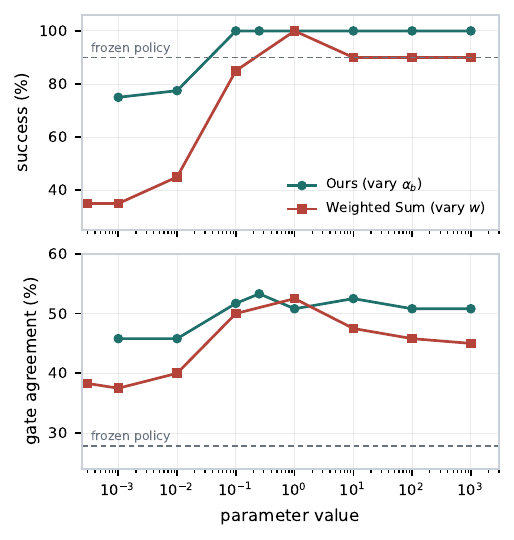}
\caption{Sensitivity of each method to its own parameter, averaged over four requested
routes ($40$ episodes per point). Our method reaches the best value of both metrics by
$\alpha_b{=}10^{-1}$ and holds it for every larger value we tried. The weighted sum
reaches
the same best values only near $w{=}1$. Success alone is misleading at large $w$: it
recovers to the frozen policy's level because the method stops following the preference,
which the
gate-agreement panel shows.}
\label{fig:d3il_sens}
\end{figure}

Both methods use the same step size and differ only in how they weight the feasibility gradient: ours through the barrier constant $\alpha_b$, the baseline through a fixed weight. We sweep each from $10^{-3}$ to $10^{3}$. The two parameters are not the same quantity, so setting them to equal values does not make the comparison fair. We instead compare the methods at equal preference compliance, asking how much success each retains at a given level of gate agreement (Fig.~\ref{fig:d3il_sens}).

We observe that the two methods reach the same best performance.
This is the expected outcome for a task with one characteristic scale, and it
is consistent with the LIBERO result. Fig.~\ref{fig:d3il_traj} shows the corresponding
paths: at their best settings the two methods produce the same behavior.


\begin{table}[t]
\caption{Success (\%) as each method's own parameter is varied.}
\label{tab:d3il_sens}
\centering
\scriptsize
\setlength{\tabcolsep}{4pt}
\begin{tabular}{lcccccccc}
\toprule
Parameter value & $0$ & $10^{-3}$ & $10^{-2}$ & $10^{-1}$ & $1$ & $10$ & $10^{2}$ & $10^{3}$ \\
\midrule
Ours ($\alpha_b$) & $75.0$ & $75.0$ & $77.5$ & $100$ & $100$ & $100$ & $100$ & $100$ \\
Weighted Sum ($w$) & $35.0$ & $35.0$ & $45.0$ & $85.0$ & $100$ & $90.0$ & $90.0$ & $90.0$ \\
\bottomrule
\end{tabular}
\tabnote{Note that
$\alpha_b{=}0$ does not remove the feasibility term: it drops the required decrease but
still cancels any update that would increase $g$.}
\end{table}

However, \emph{they differ in how much tuning it takes.}
Table~\ref{tab:d3il_sens} and
Fig.~\ref{fig:d3il_sens} sweep each method's own parameter over the same span, with the
same barrier and the same step size. Our method reaches full success for any $\alpha_b
\geq 0.1$ and keeps it all the way to $\alpha_b{=}10^{3}$; it degrades only when
$\alpha_b$ is small enough that the barrier stops asking for a decrease, and even then
it stays well above the weighted sum's worst settings. The weighted sum reaches full
success only at $w{=}1$ and loses performance on both sides: too small and it does not
enforce feasibility, too large and it holds the rod away from the gaps and hurts
success.

The reason is visible in the update rule. The barrier multiplier $\lambda$ is chosen so
that the feasibility cost decreases at a target \emph{rate}, so raising $\alpha_b$ past
what
is needed changes nothing: the multiplier cannot overshoot. A fixed weight instead adds a
fixed \emph{force} wherever the feasibility gradient is non-zero, so a weight that is too
large is as harmful as one that is too small.

\begin{table}[t]
\caption{Success ($\%$) as the steering step size increases.}
\label{tab:d3il_transfer}
\centering
\small
\begin{tabular}{lccc}
\toprule
Setting & $\eta{=}8$ & $\eta{=}16$ & $\eta{=}32$ \\
\midrule
Ours, $\alpha_b{=}0.25$ & $\mathbf{100.0\pm0.0}$ & $\mathbf{100.0\pm0.0}$ & $\mathbf{75.0\pm0.0}$ \\
W.\ Sum, $w{=}1$        & $\mathbf{100.0\pm0.0}$ & $93.3\pm4.4$  & $55.6\pm2.5$ \\
W.\ Sum, $w{=}0.1$      & $95.6\pm1.0$  & $93.3\pm1.7$  & $73.3\pm1.7$ \\
W.\ Sum, $w{=}10$       & $89.4\pm1.9$  & $77.2\pm8.2$  & $52.8\pm3.5$ \\
\bottomrule
\end{tabular}
\tabnote{Note that the best fixed weight changes with the step size; the barrier constant does not.}
\end{table}

\emph{The best weight also moves.}
Table~\ref{tab:d3il_transfer} increases the steering step size, which increases the
pressure the preference puts on feasibility, and repeats every setting three times with
independent seeds. The best fixed weight is not the same at every step size: $w{=}1$ is
best at $\eta{=}8$ but drops to $55.6\%$ at $\eta{=}32$, while $w{=}0.1$
survives the largest step size but gives up performance at the smallest. Our method uses
the same $\alpha_b$ throughout and is at or above the best fixed weight at every step
size.
Its results are also more repeatable: the standard deviation across repeats is $0.0$ at
every setting, whereas the fixed weight reaches $8.2$ points.


We do not claim a significant advantage in mean success here. The
difference we do measure reliably is in sensitivity and repeatability, not in the best
achievable score.

\subsection{Summary}
Taken together, the two manipulation experiments show that, when the cost scales are similar across a task, a fixed weight can hold the priority order as expected. The benefit of imposing the order explicitly in this regime is that it removes the
need to find that weight: any barrier constant of $0.1$ or above works, and it does not need to
be re-tuned when the steering strength changes, whereas the fixed weight has a narrow
working range that moves with the operating point. The navigation benchmark shows the
other
regime, where the cost scales vary between episodes and no single weight serves them all.


\section{Toy Example with Deeper Hierarchy}
\label{sec:toy}
\begin{figure}[t]
\centering
\includegraphics[width=\columnwidth]{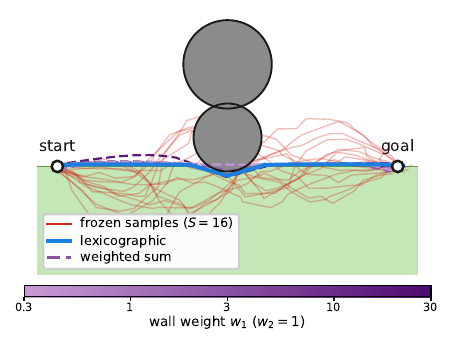}
\caption{Three-level toy example. Wall avoidance ($g_1$)
has highest priority, lawn avoidance ($g_2$) is next, and path length ($\f$)
is last. Gray discs are walls and green is lawn. Red curves are the $S{=}16$
samples from the frozen policy; blue is our selected path. Dashed purple curves
show weighted-sum results for $w_2{=}1$, with color indicating the wall weight
$w_1$. Weights $0.3$, $1$, and $3$ cross a wall, whereas $10$ and $30$ take the
lower route.}
\label{fig:toy_paths}
\end{figure}
\begin{figure}[t]
\centering
\includegraphics[width=\columnwidth]{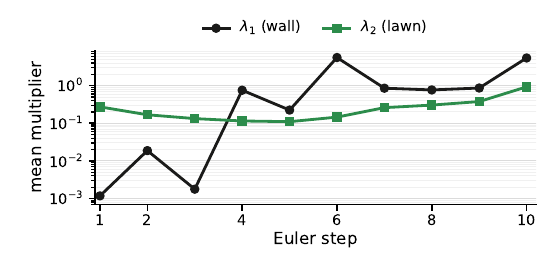}
\caption{Mean QP multipliers for the wall and lawn, averaged
over $16$ candidates and all updates at each Euler step. Their values change as
the paths change, whereas weighted-sum weights stay fixed.}
\label{fig:toy_multipliers}
\end{figure}

To show how the method handles more than two ordered costs, we train and freeze a
flow-matching policy in a simple two-dimensional task as shown in \Cref{fig:toy_paths}. At each Euler step, the proposed method steers each sample toward a shorter path while keeping the wall and
lawn costs from increasing to first order. All guided methods use the same
cascade to select the final path. 

Over $200$ episodes, $66.5\%$ of unsteered paths enter a wall by more than
$1$\si{\cm}. The proposed method has no such collisions, reaches the goal in $98.5\%$
of episodes, and travels $2.24$\si{\m} on lawn on average. The average path length is
$10.22$\si{\m}, compared to the $10.005$\si{\m} ground-truth
shortest wall-safe path. 

Only five of the $20$ weighted-sum settings avoid all wall collisions. The best
safe setting, $(w_1,w_2)=(30,3)$, has similar lawn travel to our method
($2.19$\si{\m} versus $2.24$\si{\m}), but $(30,10)$ enters a wall by more
than $1$\si{\cm} in $189/200$ episodes. Thus, a weighted sum can match our
method when tuned well, but the hierarchy avoids a search for the right
wall-to-lawn weight ratio.

\section{Conclusions}
We presented an inference-time method for steering frozen generative robot
policies under lexicographically ordered runtime costs. A dynamic-barrier
gradient update steers each sampled trajectory toward lower-priority
preferences while protecting higher-priority feasibility to first order, and a
lexicographic selection rule filters the resulting candidates one priority
level at a time. Both mechanisms act on sampled trajectories, so the same
implementation applies to diffusion and flow-matching policies without
retraining. On a last-mile delivery benchmark with conflicting traversability
and surface-preference costs, the method improves success, preference
compliance, and traversability over the frozen policy and achieves better
compliance than weighted-sum baselines tuned with the same preference
information. The same method transfers to a flow-matching manipulation policy
on LIBERO, and a controlled D3IL study shows that the dynamic barrier holds
its performance over a far wider range of settings than a fixed weight.

The main limitations are locality: steering reshapes trajectories near the
policy's own proposals, so it cannot recover a goal the frozen policy cannot
represent or supply global guidance the policy lacks. Future work includes 
steering policies with latent-space samplers where
the sample-to-workspace map is learned rather than fixed.

\bibliographystyle{IEEEtran}
\bibliography{paper/references}

\end{document}